# Prediction of Homicides in Urban Centers: A Machine Learning Approach

José Ribeiro[1,2], Lair Meneses[2], Denis Costa[2], Wando Miranda[3], Ronnie Alves[1,4]

[1] Federal University of Pará, Belém – PA, Brasil
[2] Federal Institute of Pará, Ananindeua – PA, Brasil
[3] Secretariat of Public Security and Social Defense, Belém – PA, Brasil
[4] Vale Institute of Technology, Belém – PA, Brasil

jose.ribeiro@ifpa.edu.br, lair.meneses@ifpa.edu.br,
denis.costa@ifpa.edu.br, wandomiranda@outlook.com,
ronnie.alvesr@itv.org

**Abstract.** *Relevant research has been highlighted in the computing community to develop machine learning models capable of predicting the occurrence of crimes, analyzing contexts of crimes, extracting profiles of individuals linked to crime, and analyzing crimes over time. However, models capable of predicting specific crimes, such as homicide, are not commonly found in the current literature. This research presents a machine learning model to predict homicide crimes, using a dataset that uses generic data (without study location dependencies) based on incident report records for 34 different types of crimes, along with time and space data from crime reports. Experimentally, data from the city of Belém - Pará, Brazil was used. These data were transformed to make the problem generic, enabling the replication of this model to other locations. In the research, analyses were performed with simple and robust algorithms on the created dataset. With this, statistical tests were performed with 11 different classification methods and the results are related to the prediction's occurrence and non-occurrence of homicide crimes in the month subsequent to the occurrence of other registered crimes, with 76% assertiveness for both classes of the problem, using Random Forest. Results are considered as a baseline for the proposed problem.*

**Keywords:** Prediction, Homicide, Crime, Tabular, Classification.

## 1       Introduction

New Technologies in smart cities are increasingly part of our daily lives. Several technologies are helping cities to become green, clean, organized, and safe. It is important to note that intelligent machine learning models provide part of these technological advances present in our daily lives [1].

With the large accumulation of data by institutions in the area of public security (criminological data), researchers have been able to create models based on machine learning that perform crime prediction [2] [3] [4] [5].

Computational problems in the area of criminology, such as identifying a criminal profile, exploring the context of crime, and predicting crimes, have shown interesting challenges for the computing area. These challenges have enabled the development of research focused on these themes, through different forms and perspectives [6] [7] [8] [9][10].

Criminology data is directly linked to time and space characteristics of different regions where crimes were recorded [11], which makes it inappropriate for the same machine learning model created and trained for a specific region to be used in another unknown region. Seeking to make the creation of different data exploration surveys possible, cities located in different important cities and countries around the world have made available their data related to the occurrence of crimes, such as: Ontario – Canada [12], Toronto – Canada [13], England and Wales [14], San Francisco - United States [15], and Boston - United States [16].

In the creation of a machine learning model, for regression or classification, all stages of its development are directly linked to the type and nature of data's context and the proposed problem [11]. Thus, when working with criminology data, for example in a homicide prediction model, its characteristics and development stages are unique and adapted to the specificities of time and space, linked to the context of a region. However, part of the model development strategy may be adapted and used for other regions in the same country or even in the world. For example, we can cite [2] and [5], those present similar strategies for crime type prediction, but databases from different locations.

The dynamics of how different crimes occur in a city can be explained by different theories in the area of Social Sciences, such as those related to a single person, called "Theory of Understanding the Motivations of Individual Behavior", and also from the "Theory of Associated Epidemiology", related to the study of how criminal behavior is distributed and displaced in time and space of a locality [17] [18]. The ways we seek to understand the relationship between the different crimes dealt with in this research are inspired by this second theory.

Research shows the existence of an interrelation between the number of occurrences of different types of crimes, through a comparison of historical data of specific single crimes and also through groupings of crimes, such as crime against the person (consummated murder, attempted homicide, consummated rape, attempted rape, kidnapping, etc.) and crimes against property (theft, armed robbery, theft followed by death, etc.) [19][20][21]. That is, with data referring to the number of different crimes, it becomes possible to carry out a prediction process for a crime of interest.

Data related to the general context of a crime, which may explain different aspects that motivated the occurrence of it, can assume high dimension because, in addition to time and space [11], data from social networks [22], specific personal conditions of individuals can be used [23], as well as specific contexts of a problem to be solved [24] and even climatic data from the environment [2]. However, we focus on working with time and space data of the occurrence of crimes in to predict homicides.

This research presents a way, different from that existing in the literature, regarding the transformation of criminology data that allows the creation of a dataset aimed at generically predicting homicide crimes, without dependencies on specific attributes of the city of study. To standardize the proposal of attributes used by the model (input) presented by here, an experimental study was carried out with data from occurrence bulletins in the city of Belém do Pará, from the years 2016 to 2019, which can be replicated to other locations.

From the elaboration of the proposed model, analysis, and discussions, the main contributions of this research are:

- For the machine learning area related to the problem of prediction of specific crimes. In this aspect, the data transformations performed for the creation of the proposed tabular database are different from those found in the current literature, since they confront the numbers of different types of crimes (independent variables) with the prediction of the occurrence or non-occurrence of homicide crimes in the near future. Considering that this dataset was developed to minimize the dependence on characteristic or specific attributes of the study region so that it can be more easily replicated to other regions, since there is a description of the generic attributes used (inputs) of the models in this work;

- For the community that works with criminology data, as this study offers a baseline of performance of machine learning models of different complexities related to the problem of predicting homicide crimes, which can be very well adapted and replicated in other cities, that have criminology data similar to those used by this research.

## 2     Related work

This research did not find works related to the prediction of homicide crimes using data similar to those worked here. However, some researches approach ways of predicting the types of crimes (in general, not just a specific one), something similar but still different from what is presented in this study.

In this sense, machine learning surveys using crime data are important both for the computing community and for society in general. With this, research such as:

- "Crime Type Prediction". Description: Research objective is to identify the spatial, temporal, weather, and event features most often associated with specific crime types, using machine learning models based on different algorithms: Logistic Regression, Random Forest, Decision tree, and XGBoost. The data used were Chicago Crime Data, Historical Hourly Weather Data 2012–2017, all referring to the United States [2];

- "San Francisco Crime Classification". Description: Proposal of classificatory machine learning models capable of predicting the type of crime that may occur in the city of San Francisco in the United States, from the time and place of the crime in question. The tested algorithms are: Logistic Regression, Naive Bayes, and Random Forest [25];

- "A Prediction Model for Criminal Levels Specialized in Brazilian Cities". Description: This paper proposes a model of data mining, predicting criminal levels in urban geographic areas. The model was proposed to work using Brazilian data (from the city of Fortaleza – CE) between 2007 and 2008, specifically criminal and socio-economic ones. Several algorithms were tested, but the best results were collected with neural networks [3];

- "Addressing Crime Situation Forecasting Task with Temporal Graph Convolutional Neural Network Approach". Description: Article on a proposed machine learning model based on Graph Convolutional Neural Network and Recurrent Neural Network to capture the spatial and temporal dynamics of crime statistics recorded in the city of San Francisco - California, USA, from 2003 to 2018 [10];

- "Crime Data Analysis and Prediction of Perpetrator Identity Using Machine Learning Approach". Description: A complete article on analyzing and predicting the profile of perpetrators of crimes using machine learning. In this study, homicide data from the city of San Francisco (United States) from 1981 to 2014 were used [26];

- "Crime Pattern Detection, Analysis & Prediction". Description: A complete article on analyzing crime data by detecting patterns and predictions. The data used in this research refer to six different categories of crimes registered in the interval of 14 years (2001-2014), referring to the city of Thane (India) [9];

- "Predictive Policing Software - PredPol". Description: Software for police use, focused on the monitoring and analysis of several variables in a micro-region, enabling the prediction of the probability of occurrence of specific crimes with a location and time suggested by the tool. This tool is a success story of the application of intelligent algorithms to criminology data, as it is currently used by security institutions in countries such as the United States of America. Details about the data analyzed by this tool to make predictions are omitted by the developers. [27];

From reading the works cited above, there are different approaches applied to problems in the area of criminology, highlighting methodological strategies based on algorithms for prediction, time series analysis, use of spatial data as input of the model, non-use personal data of individuals, and use of data from police reports from different cities around the world.

It is important to highlight that in all studies cited machine learning models, directly or indirectly, use information from crimes related to time and space, as well as the research described here.

Another important observation is that in the studies [2][25] [3][10][26] [9] the dataset created and used by machine learning models have attributes of a specific nature (dependency) of the city of study to which the research is linked, making the models less replicable to other locations.

In this way, this research proposes a machine learning model, based on data time, space, and different amounts of crimes registered in the city of study, without presenting significant dependencies on specific variables of the locality in question.

## 3 Crime Understanding

Explaining how crimes arise and are related to each other in an urban center is not an easy task, as there is a large set of data that can be explored in to seek a coherent explanation of what leads individuals to commit crimes [11].

Researches highlight the existence of correlations between the high occurrences of crimes and socioeconomic variables in some regions, especially when the group evaluated in question are people in social vulnerability [28] [29]. This being one of the ways to explain how the process of disseminating crimes occurs in urbanized cities. However, it has not yet been used in the current maturity of this work.

The existence of older studies and theories in the area of Social Sciences, which explain social behavior linked to crimes, called Associated Epidemiology - AE, which studies aspects of how criminal behavior is distributed in space and displaces over time [17].

More recent research [18] explained how one of the aspects referring to AE can be described through the Theory of Social Disorganization, which is defined as a systemic approach, in which the main focus is the local communities, understanding these as a system of networks of formal and informal associations regarding friendships, relatives, jobs, cultures, economies and even crimes that influence the human living in society [30].

The City of Belém, capital of the state of Pará in Brazil, the scenario of this study, is a city with 1,393,399 inhabitants, according to its last survey carried out in 2010, and has a human development index of 0.746 [31]. This city has institutions in the security area that carry out various actions aimed at tackling crime. Even so, it still presented the third worst homicide rate among all Brazilian captains (74.3) in a study carried out in 2017 [32]. In this way, the data from the records of occurrence bulletins in this city prove to be interesting sources of information for the study carried out here.

Inspired by the above, this work applied a series of cleaning, transformation, and pre-processing steps to the database of police reports, allowing the generalization (by an algorithm) of different associations between different crimes registered in the city of study, facing the challenge of predicting homicide crimes in a predefined period of time.

## 4 Machine Learning Approach

### 4.1 The data

The data used in this research were provided by the Assistant Secretariat of Intelligence and Criminal Analysis - SIAC of the state of Pará, Brazil. Such data refer to the police reports registered during the years 2016 to 2018 in the city of Belém - Pará.

The raw data can be characterized as transactional tabular data containing information such as crime's id, occurrence date, registration date, time, time, type, description, cite, neighborhood, unit from the register, and others 31 administrative context attributes.

In terms of size, the database has 41 attributes per 507,065 instances. Where each instance represents a police report registered in the city. Only 4 of the 41 attributes mentioned above were used to create the new database, the main reason being related to the lack of filling in of the other data, as well as the non-direct relationship with the crime context. The 4 attributes in question are crime's occurrence date, type, municipality, and neighborhood.

The database has records related to more than 500 different types of crimes. Highlighting the crimes of theft, damage in traffic, threat, other atypical facts, bodily injury, embezzlement, and injury, as the eight most common crimes in the base, nomenclatures defined by the Civil and Criminal codes of Brazil [33][34].

Pre-processing and specific data transformations made it possible to analyze how different crimes are dynamically related to homicides in the city of study. Because the occurrence of specific crimes is related to a context of the conflict between individuals and one crime may influence homicides [24][17][30].

### 4.2    Pre-process

The pre-processing procedures applied in the database of the proposed model are presented in the next paragraphs.

Attributes exclusion: 9 Sparse attributes (with unregistered data) and id; 21 Attributes not directly related to the crime context; 2 Attributes related to personal data of registered individuals; 2 Attributes related to the location (street) of the crime that occurred due to inconsistency; And 2 attributes of crime's georeferencing (latitude and longitude) due to inconsistency.

Exclusion of records related to occurrences in neighborhoods located on small islands or in rural areas due to high inconsistency; Police reports instances considered non-crimes; Duplicated records (since a crime can be registered in more than one police report, in this case by different people);

We also make the removal of special characters (such as: @#$%ˆ&*áíóúç?!ᵒᵃ•§∞¢£™¡) and the consolidation of neighborhoods in the city of study.

In this preprocessing and cleaning, there was a decrease from 507,065 instances to 453,932. Attributing such data loss to higher quality and consolidation.

Then, we do the transformation of the database from tabular transactional to tabular based on time and space with minimum granularity equal to a month. At this stage, only neighborhoods that had records of crimes in all months of the years analyzed were considered, aiming to minimize loss of information in specific neighborhoods.

Among the preprocessing procedures listed above, more characteristics of the transformation of the tabular transactional database to a tabular based in time and space database were inspired by Online Analytical Processing – OLAP [35].

This transformation was necessary because the objective model needed to perform the prediction of crimes according to time in months. In this way, the tabular transactional database was transformed into a new tabular database, considering the year of the crime, the month of the crime, and the neighborhood of the crime concerning the numbers of each of the registered crimes, Table 1.

**Table 1.** Illustration of tabular dataset construction.

| year* | month | neighborhood* | threat count | theft count | homicide count | ... | Class |
|---|---|---|---|---|---|---|---|
| 2016 | 1 | 1 | 3 | 5 | 5 | ... | 1 |
| 2016 | 1 | 2 | 5 | 7 | 0 | ... | 0 |
| 2016 | 1 | 3 | 4 | 20 | 4 | .. | 1 |
| .. | .. | .. | .. | .. | ... | .. | .. |
| 2016 | 2 | 1 | 5 | 40 | 5 | ... | 1 |
| 2017 | 2 | 2 | 1 | 39 | 4 | ... | 1 |

Note in Table 1, the Class attribute (with values between 0 and 1) defines whether or not homicide occurred in the month following the date of crime instance, taking into account the year, month, and neighborhood of the instance.

In table 1, the year and neighborhood columns appear to facilitate the understanding of the tabular dataset's construction. However, as presented in the pre-processing, these attributes are not used as model inputs. In other words, only attributes related to time (month) together with the various amounts of other crimes are used as input for the machine learning models developed.

This research emphasizes that it carried out preliminary analyses of the data in order to verify the possibility of working with data granularity equal to day and week, yet no strong correlations were identified between the model's input attributes and the prediction class (homicide in the next day or homicide the next week) in both cases.

To perform data fairness, it was decided to leave the neighborhood attribute outside the algorithm entries, as it was identified that this attribute was able to map the objective class with an accuracy close to 80% for some specific neighborhoods.

In summary, the attributes year and month were used to organize the data, as well as the spatial attribute neighborhood. However, both year and neighborhood (marked with a '*' in Table 1) are considered meta-attributes by this research and only participate in data modeling (not being passed on as input to the algorithms). The remaining data, except for the class, refer to the count of specific crimes (in years, months, and specific neighborhoods).

To better explain the treatment, follows the example: in Table 1, line 1 shows a record of the year 2016, month 1 (one), *neighborhood* 1 (one) and *threat count* 3 (three), *theft count* 5 (five), *homicide count* 5 (five) and *Class* 1 (one). The *Class* had a value 1 (one) because there was a homicide in the following month (month 2) of this record in the same *neighborhood* 1 (one), as can be seen in the penultimate line of same figure, specifically in the column *homicide count* 5 (five). If this last cited column had a value of 0 (zero), the class in question would be 0 (zero). Thus, the class only presented a value equal to 1 (one) because the *homicide count* was greater than 0 (zero).

The new tabular database class was processed to become binary, now showing values of 0 (zero) or 1 (one). Being 0 (zero) the absence of homicide and 1 (one) the existence of homicide, taking into account a specific year, month, and neighborhood. The balance between classes in the dataset was: 970 instances for class 0 (non-homicide) and 1,034 for class 1 (homicide).

Nonetheless, the new database suffered a significant dimensionality reduction, with 2,004 instances, 36 attributes (34 quantitative of different crimes, 1 ordinal attribute referring to the month), and 1 binary class (0 and 1). All 34 attributes of crime numbers went through min-max normalization, obtaining values between 0 and 1 [36]. The month attribute has been converted to an entire numeric ranging from 1 to 12 (equivalent to each year's months).

This research did not carry out any process of reducing the dimensionality by automatic methods such as feature selection, considering it unnecessary at this moment, since it is desired to obtain the maximum information from a crime context.

The 34 attributes of crimes used in these studies are: bodily injury, threat, assault, injury, theft, traffic injury, traffic damage, defamation, homicide, abandonment of the home, vicinal conflicts, marital conflicts, escape from home, rape vulnerable, other atypical facts, vehicle theft, embezzlement, damage, civil damage, slander, family conflicts, drug trafficking, aggression/fight, misappropriation, physical aggression, reception, rape, the disappearance of people, attempted murder, pollution sound, other frauds, disobedience, contempt, and disturbances of tranquility.

As shown, the data used in this research are the most reliable as possible, since they were provided by the public security institution of the study city. However, the presence (even if minimal) of noise in the data must be admitted, with this it is emphasized that the models developed in this research are tolerant of data errors.

As seen, the data used as inputs for machine learning models are completely generic, since they are made up of the month variable (1 variable) along with the different numbers of crimes that occurred in neighborhoods in a specific month (34 variables), which makes this methodology of using criminology data for the prediction of homicides easily replicable to other cities that have the same data. More details about the dataset and analysis of this study Git: >>> https://github.com/josesousaribeiro/Pred2Town.

### 4.3 Analyzed algorithms

After the database was cleaned, consolidated, pre-processed, transformed, and with dimensionality reduction previously presented, this research carried out a series of tests with algorithms of different potentials: lazy learners (represented by K-Nearest Neighbors – KNN [37]), eager learners (represented by Support Vector Machine – SVM [38], Decision Tree – DT [39], Neural Network – NN [40], Naive Bayes – NB [41], Logistic Regression – LR [42]) and ensemble learners (represented by Gradient Boosting – GB [43], Random Forest – RF [44], Extreme Gradient Boosting – XGB [45], Light Gradient Boosting - LGBM [46], and CatBoosting – CB [47]), all implemented in python.

As seen above, the idea is to carry out the process of creating machine learning models not only using robust algorithms, such as the cases of ensemble learners, but also simpler and faster learning algorithms, as is the case with lazy and eager learners, seeking to evaluate which of the algorithms can better exploit the database.

An important observation is that only Decision Tree, K-Nearest Neighbors, Naive Bayes, and Logistic regression algorithms are considered transparent algorithms (with high explicability), a characteristic desired in the context of the proposed homicide prediction problem. The other algorithms are considered black boxes, so they end up not being self-explanatory.

The database was divided into training (70%) and test (30%) using stratification by the objective class, being careful so that records from all neighborhoods and all years existed (in similar proportions) in both training data and test data.

The tuning process, Table 2, was performed using the Grid Search [48] based on cross-validation with folds size equal to 7, using the variation of common parameters among the tested algorithms, as well as exclusive parameters of each algorithm, to promote equality of variation of parameters (as far as possible), without minimizing differentials specific to each algorithm. This was performed using only the training data to carry out this process, aiming at a fairer comparison between models.

**Table 2.** Tuning process description.

| Model | Range of parameters | Best parameters found |
|---|---|---|
| CB | Learning rate: [0.1, 0.5]; Depth: [1, 6, 12]; Iterations: [10, 100, 200]; Grow policy: ['SymmetricTree', 'Depthwise', 'Lossguide']; Bagging temperature: [0, 0.5, 1]. | Learning rate: 0.1; Iterations: 200; Grow policy: 'SymmetricTree'; Bagging temperature: 0. |
| DT | Min samples leaf : [1, 10, 20, 40]; Max depth: [1, 6, 12]; Criterion: ['gini','entropy']; Splitter: ['best', 'random']; Min samples split: [2, 5, 15, 20, 30]. | Min samples leaf : 40; Criterion: 'entropy'; Splitter: 'random'; Min samples split: 2. |
| GB | Max d.: [1, 6, 12]; N. estimators: [10, 100, 200]; Min samples leaf: [1, 10, 20, 40]; Learning r.: [0.1, 0.5]; Loss: ['deviance', 'exponential']; Criterion: ['friedman_mse', 'mse', 'mae']; Max f.: ['sqrt', 'log2']. | Max depth: 12; N. estimators: 100; Min samples leaf: 40; Learning rate: 0.1; Loss: exponential; Criterion: 'mae'; Max features: 'sqrt'. |
| LGBM | Learning r.: [0.1, 0.5]; Max d.: [1, 6, 12]; Bootstrap: [True, False]; N. estimators: [10, 100, 200]; Min data in leaf: [1, 10, 20, 40]; Boosting t.: ['gbdt','dart','goss','rf']; Num. l.: [31,100,200]. | Learning rate: 0.1; Max depth: 1; Bootstrap: True; N. estimators: 100; Min data in leaf: 40; Boosting type: 'goss'. |
| KNN | Leaf size:[1, 10, 20, 40]; Algorithm:['ball_tree', 'kd_tree', 'brute']; Metric: ['str', 'callable','minkowski']; N._neighbors:[2,4,6,8,10,12,14,16]. | Leaf size: 1; Algorithm: 'ball_tree'; Metric: 'minkowski'; N._neighbors: 8. |
| LR | Solver: ['newton-cg', 'lbfgs', 'liblinear', 'sag', 'saga']; Penalty: ['l1', 'l2']; C:[0.001,0.008,0.05,0.09,0.1]; Max iter.: [50, 200, 400, 500,600]. | Solver: 'sag'; Penalty: 'l2'; C: 0.1; Max iter.: 50. |
| NB | Var. smoothing: [1e-5, 1e-7, 1e-9, 1e-10,1e-12]. | Var. smoothing 1e-5: |
| NN | Learning r.: ['constant', 'invscaling', 'adaptive']; Solver: ['lbfgs', 'sgd', 'adam']; Activation: ['identity', 'logistic', 'tanh', 'relu']; Max iter.: [200,300,400]; Alpha: [0.0001,0.0003]; hidden layer sizes: [1,2,3,4,5]. | Learning rate: 'invscaling'; Solver: 'adam'; Activation: 'tanh'; Max iter.: 300; hidden layer sizes: 3. |
| RF | Max depth: [1, 6, 12]; Bootstrap: [True, False]; N. estimators: [10, 100, 200]; Min samples leaf: [1, 10, 20, 40]; CCP alpha: [0.0, 0.4]; Criterion: ['gini', 'entropy']; Max features: ['sqrt', 'log2']. | Max depth: 12; Bootstrap: True; N. estimators: 100; Min samples leaf: 1; CCP alpha: 0.0; Criterion: 'gini'; Max features: 'log2'. |
| S V M | C: [0.001, 0.01, 0.1, 1, 10]; Kernel: ['linear', 'poly', 'sigmoid']; Shrinking: [True, False]; Degree: [1,2,3,4,5]. | C: 10; Kernel: 'poly'; Shrinking: True; Degree: 1. |
| X G B | Max d.: [1, 6, 12]; N. estimators: [10, 100, 200]; Min s. le.: [1, 10, 20, 40]; Booster: ['gbtree', 'gblinear', 'dart']; Sampling m.: ['uniform', 'gradient_based']; Tree m.: ['exact','approx','hist']. | Max depth: 1; N. estimators: 200; Min samples leaf: 1; Booster: 'gbtree'; Sampling method: 'uniform'; Tree method: 'approx'. |

Table 2 presents all the best parameters found from the execution of the grid search process based on cross-validation with folds size equal to 7, and the metric used to measure the performance of each fold execution was the Area Under ROC – AUC [49]. It was decided to use cross-validation at this stage of creation to identify the most stable machine learning models in the face of data as input.

We chose to use the AUC evaluation metric because it takes into account the successes and errors identified in both classes (1 and 0) of the problem in question. Thus, the AUC measures both successes and errors of homicides and non-homicides that occurred, a characteristic that is desirable given the nature of the problem — Since predicting a homicide is just as important as predicting a non-homicide.

## 5　Discussion

The tests performed with the 11 algorithms were divided into two moments: A) Performance Analysis: Comparison of performances based on Accuracy - ACC and Confusion Matrices; B) Statistical Analysis: based on the *Friedman test* and score AUC.

### 5.1　Performance Analysis

Note, accuracy was chosen to measure the correctness, in this stage of the tests, since it considers both true positives and true negatives in the metric calculation. That is, it considers the two classes of the problem to indicate the best performance.

Table 3 shows the results of the base tests with each of the listed algorithms (ordered by ACC). As it can be noted that the algorithms had results with accuracy values fluctuating between 0.69 to 0.76. The best model (RF), obtained accuracy equal to 0.76, accompanied by the LGBM algorithm with 0.75 and XGB with 0.75, this is a minimal difference and although it exists, it is not considered relevant.

**Table 3.** Accuracy by Model.

| MODEL | ACC | MODEL | ACC | MODEL | ACC |
|---|---|---|---|---|---|
| RF | 0.76 | LR | 0.74 | DT | 0.71 |
| LGBM | 0.75 | SVM | 0.74 | NB | 0.69 |
| XGB | 0.75 | CB | 0.74 | KNN | 0.69 |
| NN | 0.74 | GB | 0.72 | - | - |

Considering the values presented in Table 3, there is difficulty in identifying which model is better than the other, since the accuracy values are very close to each other. For even checking, the confusion matrix of all algorithms is presented in Table 4.

In Table 4, it can be noted that the three highest accuracies were found in algorithms that balanced both class 0 and class 1 (RF, LGBM, and XGB algorithms).

The NN, LR, SVM, and DT algorithms showed considerable accuracy, but they tend to classify class 0 better than class 1. What is not desirable for research, since it is understood as important for both classes of the problem, Table 4.

**Table 4.** COMPARISON OF CONFUSION MATRICES.

|   | RF | | LGBM | | XGB | | NN | | LR | | SVM | |
|---|---|---|---|---|---|---|---|---|---|---|---|---|
|   | 0 | 1 | 0 | 1 | 0 | 1 | 0 | 1 | 0 | 1 | 0 | 1 |
| 0 | **76%** | 24% | **76%** | 24% | **75%** | 25% | **80%** | 20% | **82%** | 18% | **85%** | 15% |
| 1 | 24% | **76%** | 25% | **75%** | 24% | **76%** | 31% | **69%** | 33% | **67%** | 36% | **64%** |
|   | CB | | GB | | DT | | NB | | KNN | | Note: The test class has 602 instances (311 of class 1 and 291 of class 0); | |
|   | 0 | 1 | 0 | 1 | 0 | 1 | 0 | 1 | 0 | 1 | | |
| 0 | **73%** | 27% | **74%** | 26% | **78%** | 22% | **84%** | 16% | **89%** | 11% | | |
| 1 | 26% | **74%** | 30% | **70%** | 34% | **66%** | 46% | **54%** | 50% | **50%** | | |

The results found by CB and GB, present relevant accuracy, but with slightly lower values than those presented by RF, LGBM, and XGB.

Still, concerning Table 4, NB and KNN were the algorithms that showed the greatest successes in class 0, but also the greatest errors in class 1.

Based on the analyses presented in tables 3 and 4, the RF, LGBM, and XGB algorithms are considered the best classifiers for the homicide prediction problem. However, to identify the significance of the differences between the models analyzed, statistical analyses are performed below.

### 5.2 Statistical Analysis

A statistical analysis of the 11 machine learning models was carried out to identify two main aspects of the created models: stability and statistical significance.

To assess the stability of the models, cross-validation runs were performed with fold = 7 for each model. After this execution, each model analyzes its outputs analyzed by the AUC metric, and, finally, a graph of Kernel Density Estimate (Gaussian with bandwidth 0.6) of executions was created in order to identify the AUC value for each of the 7 folds tested. This analysis is shown in Figure, 1.

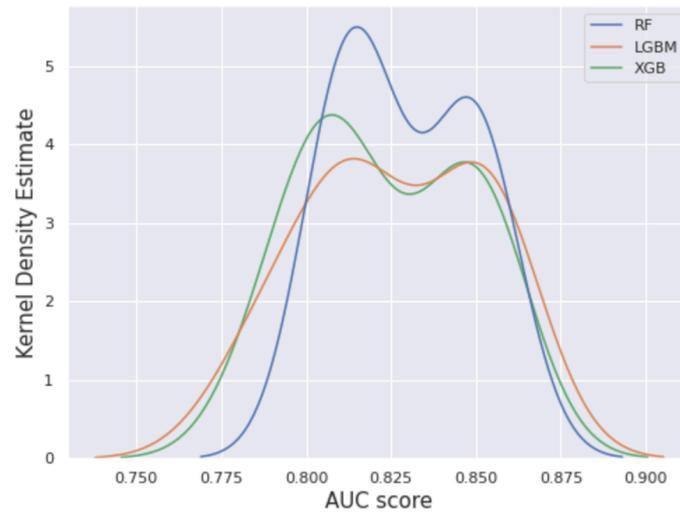

**Fig. 1.** Density of AUC scores for execution of cross-validation with fold's size equal to 7 for RF, LGBM, and XGB.

Analyzing Figure 1, it can be noted that the three algorithms present similar stability since for each of the 7 executions a similar performance variation of the algorithm between the models was identified, with values between 0.775 to 0.875 of AUC.

Still, for Figure 1, it can be verified that the RF algorithm showed greater stability when compared to LGBM and XGB since RF presented slightly higher density (higher values on the y-axis) and concentrated (interval of the x-axis) than the LGBM and XGB models.

To analyze the significance of the different classificatory results found by each algorithm, in the cross-validation runs with size 7 folds, mentioned above, statistical analyses were performed based on the Friedman test to compare each tested algorithm pair.

In the Friedman test, only p-value values <= 0.05, found for each pair of analyzed algorithms, are considered statistically significant. With this, it was possible to construct Figure 2.

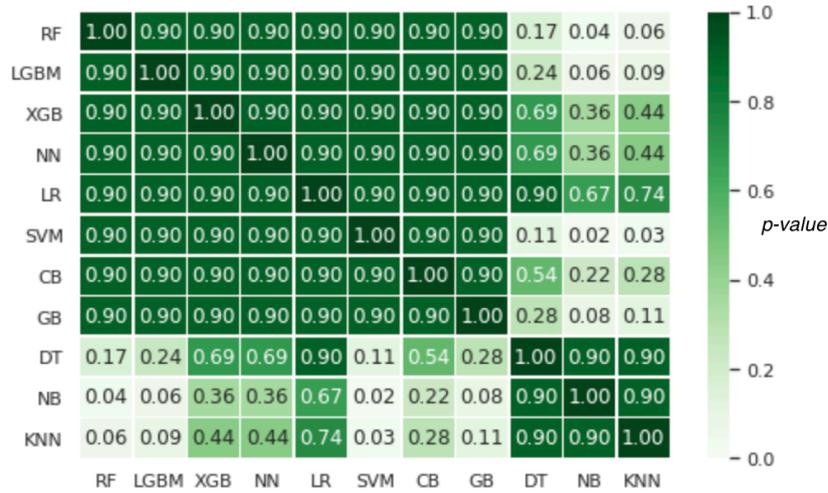

**Fig. 2.** Result of the Friedman test for all pairs of algorithms. Values of p-value <= 0.05 mean a statistically significant difference between two algorithms (row and column).

Based on the previous analyses, which point to the RF model as the best classified, and confronting them with the results presented in Figure 2, related to the summary matrix of the Friedman test, it can be verified RF may present the best results in terms of performance, but these results are not statistically significant when compared to the results of LGBM, XGBM, NN, LR, SVM, CB, GB, DT, and KNN.

However, given the context of the homicide prediction problem, which presents significant sensitivity and the need for correctness (high performance) of the model. This study recognizes the results of the Random Forest algorithm as the best presented in this case.

In order to present the performance of the RF algorithm in a contextualized way with the space of the study city, in Figure 3, a visualization layer created in the model's output is presented, which distributes each tested instance according to the neighborhood it belongs to. In this way, it is possible to have an understanding of which are the neighborhoods that the algorithm hits the most and in which it misses the most. Note that even without an algorithm receiving the neighborhood attribute as data entry, it manages to learn the patterns of the occurrence of crimes and thus can predict the occurrence of homicide.

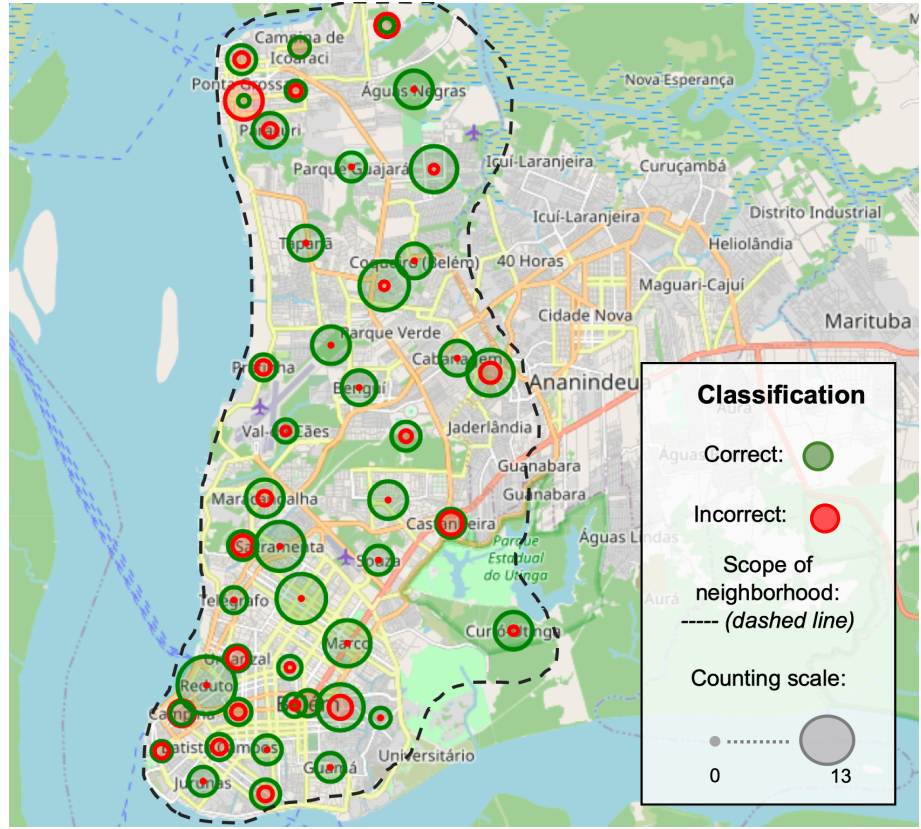

**Fig. 3.** Visualization of the Random Forest algorithm outputs, over the map of the central region of Belém, Pará - Brazil. In most neighborhoods there is a predominance of correctness of the model (green circles larger than red circles).

This research understands that in neighborhoods where the algorithm errs more frequently, red circles in Figure 3, may be related to factors such as loss of information, or even noisy data and data registration problems, however, such aspects and discourses transcend the boundaries of this work.

## 6    Conclusion

This research pre-processed, transformed, and used data from bulletins of different crimes for the creation of a prediction model of homicide crimes in a different way from the literature, being proven feasible and fair. Since the proposal highlights the non-use of data from local contexts in the analyzed spatial region and does not use individual demographic data of people, avoiding the creation of social stereotypes.

Therefore, it became feasible to use the different amounts of crimes (together with the variable month) as input to machine learning models in an attempt to predict the occurrence of homicide crimes in the following month, emphasizing that this is a set of generic data for the problem in question. It should be noted that this same proposal can be replicated to other cities that have such numbers of crimes recorded in data.

Concerning the tests performed, 1 highlight among the 11 models created was noticed, which was the Random Forest algorithm, since this model presented the best accuracy value, higher percentages of correct answers in the two objective classes, greater stability in terms of performance (Cross-validation with AUC), despite not having statistically significant differences when compared with 9 other models presented. However, this study considers the results found as a baseline for the communities of researchers working with criminal data.

In areas of knowledge, sensitive as the area of criminology, working with machine learning models aimed at prediction is new and has been gaining more space in society, but such models must have high performances and explanations. In this way, when adopting an ensemble machine learning model aiming at its performance, important aspects are lost regarding the explicability of predictions that could justify which crimes are directly related to specific instances of homicides.

The next steps of this research are intended to test computer techniques of Explainable Artificial Intelligence - XAI capable of carrying out the process of opening the black box, and thus seek to explain how and what are the different types of crimes that enabled the prediction of homicides, as presented here.

## Acknowledgment

This research acknowledges the collaboration and cooperation of the institutions involved in the development process of the presented analysis: Federal Institute of Pará, Secretariat of Public Security and Social Defense, and Vale Institute of Technology. Together with the institutions involved in making this publication possible: Federal University of Pará, and Cosmopolita Faculty.